# High-throughput Biomedical Relation Extraction for Semi-Structured Web Articles Empowered by Large Language Models

## Authors:

Songchi Zhou, BS[1,2]
Sheng Yu, PhD[1,2,*]

## Affiliation of the authors:

[1]Center for Statistical Science, Tsinghua University, Beijing, China
[2]Department of Industrial Engineering, Tsinghua University, Beijing, China

## Corresponding author:

Sheng Yu
Weiqinglou 209, Tsinghua University
Beijing, 100084, China
E-mail: syu@tsinghua.edu.cn

# Abstract

**Objective:** To develop a high-throughput biomedical relation extraction system that takes advantage of the large language models'(LLMs) reading comprehension ability and biomedical world knowledge in a scalable and evidential manner.

**Methods:** We formulate the relation extraction task as binary classifications for large language models. Specifically, LLMs make the decision based on the external corpus and its world knowledge, giving the reason for the judgment for factual verification. This method is tailored for semi-structured web articles, wherein we designate the main title as the tail entity and explicitly incorporate it into the context, and the potential head entities are matched based on a biomedical thesaurus. Moreover, lengthy contents are sliced into text chunks, embedded, and retrieved with additional embedding models.

**Results:** Using an open-source LLM, we extracted 248659 relation triplets of three distinct relation types from three reputable biomedical websites. To assess the efficacy of the basic pipeline employed for biomedical relation extraction, we curated a benchmark dataset annotated by a medical expert. Evaluation results indicate that the pipeline exhibits performance comparable to that of GPT-4. Case studies further illuminate challenges faced by contemporary LLMs in the context of biomedical relation extraction for semi-structured web articles.

**Conclusion:** The proposed method has demonstrated its effectiveness in leveraging the strengths of LLMs for high-throughput biomedical relation extraction. Its adaptability is evident, as it can be seamlessly extended to diverse semi-structured biomedical websites, facilitating the extraction of various types of biomedical relations with ease.

# Introduction

Biomedical relation extraction (BioRE) is a fundamental research topic in biomedical natural language processing (BioNLP), offering the potential to construct expansive biomedical knowledge graphs (KGs) [1] and enhance clinical practices [2]. Abundant biomedical websites, such as Medscape and MedlinePlus, house extensive essential biomedical knowledge and are ideal sources for relation extraction. We aim to develop an automated mechanism to extract relations from those websites in a high-throughput manner, i.e., without significant human effort in sample annotation or model training for each relation type or data source. A common format of articles on those websites is that the content is intricately tied to the main title (i.e. a disease), so the main title can be naturally designated as the tail entity, and potential head entities are terms found in the narrative text on the webpage. In particular, we are interested in leveraging information in addition to the text to improve accuracy, such as the HTML structure, the section hierarchy, and entity semantic types, which we refer to as semi-structured input. The desired outcome in relation extraction is the extraction of a relation triplet in the form of (head entity, relation, tail entity). A naive approach to automate relation extraction involves using pattern-based methods like regular expressions or finite state machines [3]. However, patterns are oftentimes too complicated to recognize for these simple pipelines, resulting in the extraction of only a limited number of relation triplets. Notably, the term serving as the main title may not appear in every sentence, considering that all the text within the page inherently describes the main title.

Most of the previous works [4-9] have approached biomedical relation extraction as a generic classification task using established benchmarks (e.g. BioRED [4], CDR [10], CHEMPROT [11]). In these cases, the context contains both the tail entity and the head entity, focusing on potential relation triplets at the sentence or document level. These methods are confined to handling purely narrative text, such as PubMed abstracts or electronic health record (EHR) descriptions, and are limited in their ability to extract a comprehensive range of semantic relation types.

In addition, conventionally, training relation extraction models for specific data sources necessitates sufficient labeled data. However, this proves challenging in real-world scenarios due to the considerable cost associated with human labor. Consequently, distant supervision methods [12, 13] are proposed to mitigate the problem, using external databases to label the corpus so as to construct a training set for machine learning methods. However, these methods suffer from problems including partial positive annotation, difficulty in generating hard negative samples, and the false positive problem since the context containing the entities does not necessarily state a corresponding relation in the database. Although Yoon et al. [12] proposed to use extra human-annotated data to refine the initial model, the time and financial expenses associated with human annotation may render it impractical to train a model with good performance, particularly in resource-constrained settings.

To the best of our knowledge, no studies have specifically delved into biomedical relation extraction from open semi-structured biomedical web articles. Some research introduces the problem of open information extraction (OpenIE) on general semi-structured websites, trying to solve it with distant supervision methods or using the visual layout information to help extract relation triplets [14, 15]. Additionally, methods directly modeled on the DOM tree structures or HTML text [16, 17] offer valuable insights into information extraction from semi-structured websites. However, it's worth noting that these methods primarily focus on simple web structures and generic texts, whereas biomedical relation extraction demands a more nuanced understanding of knowledge-intensive contexts.

In this work, the curated data differ from previous benchmarks for biomedical relation extraction in that: (1) The webpage HTML is preprocessed to a structure that encapsulates raw narrative texts while preserving the distinctive characteristics of semi-structured web articles such as the section hierarchy. (2) As commonly seen in real-world scenarios, the context within each text sample may not furnish sufficient information to discern the relation. Correctly classifying the relation may require world knowledge about the involved entities. (3) Finally, we aim for zero-shot relation extraction, as biomedical web articles exhibit a diverse array of semantic relations, making it impractical to construct numerous corresponding training sets for each relation due to associated costs.

Large Language Models (LLMs) have emerged as a powerful generative artificial intelligence, surpassing their predecessors in a series of natural language processing tasks [18]. In particular, ChatGPT, introduced by OpenAI in November 2022, presents a phenomenal performance in adhering to human instructions and demonstrates an extensive knowledge base when responding to free-text queries. From then on, a proliferation of both closed-source and open-source LLMs has significantly transformed the landscape of natural language processing. In the realm of biomedicine, LLMs have demonstrated substantial potential across numerous applications [19-21]. However, there are also drawbacks of LLMs [22], such as hallucination, high pre-training costs, restricted context length, etc. In clinical practice, for instance, leveraging ChatGPT to optimize clinical decision support (CDS) proves to be a judicious strategy for complementing CDS alerts [23]. Nevertheless, the persistence of hallucinations and the risk of misleading outputs from LLMs are concerning, particularly in real clinical settings where such errors are essentially unacceptable. Thus, employing a qualified text database to facilitate LLM responses proves instrumental in mitigating associated problems. Notably, Retrieval-Augmented Generation (RAG) introduces new dimensions to LLMs by incorporating relevant texts from an external database and combining the context with user queries, thus remarkably reducing the hallucination and keeping the language models up to date for accurate question answering [24-26].

In this work, we regard the preprocessed web articles as the external data source and use the LLMs as the frozen engine (i.e. no update for model parameters) for high-throughput relation extraction in a question-answering style. Our approach involves the initial extraction of all biomedical terms as potential head entities from the webpage text. We then streamline the relation extraction process into a binary classification task. Specifically, LLMs assess each pair comprising the matched term and the main title, making decisions based on binary classification. Subsequently, the models provide a rationale for each decision—a crucial aspect for verifying the correctness of provenance and minimizing the risk of hallucination in the extraction process.

In summary, our contributions are as follows:

- We propose a first-of-its-kind framework that utilizes the LLM to extract relation triplets from biomedical texts of semi-structured biomedical websites in a scalable and evidential manner. Our framework requires no data labeling and model training. It capitalizes on the reading comprehension capabilities and extensive world knowledge inherent in LLMs, mitigating the risk of hallucination.

- We extract a total of 248659 biomedical relation triplets of three kinds of relation types as demonstration of usefulness and scalability. More diverse types of relation triplets can be extracted with minimal adjustments to the current framework.

- We validate the extraction correctness by comparing with expert annotations and

show that the framework with open source LLMs can achieve the same level of F1 score (0.882) as top tier commercial models such as GPT-4.

## Materials and methods

### Data collection

Three representative biomedical websites are used to experiment: Medscape[1], MedlinePlus[2], and MSD Manual[3]. The raw HTML texts are processed to maintain hierarchical section-level information to help the LLMs better understand the context. Web articles from each website's disease categorization are extracted. For Medscape, we retain the four kinds of sub-pages (i.e. Overview, Presentation, Workup, and Treatment) of each main page. There are list structure data in the web articles, for which we use |1|, |2|, and |3| to identify the three-level structure and use || for the other three websites. All the HTML data were accessed on September 29, 2023. We have obtained written permissions from the respective authorities for using these material for researh. Biomedical Informatics Ontology System (BIOS) [27] version 2.2 is utilized as the thesaurus to provide biomedical terms along with their semantic types.

---

[1] Medscape Drugs & Diseases (Reference). Published by WebMD, LLC. Online at: https://emedicine.medscape.com/.
[2] A.D.A.M. Medical Encyclopedia [Internet]. Johns Creek (GA): Ebix, Inc., A.D.A.M.; Available from: https://medlineplus.gov/ency/
[3] From the MSD Manual Professional Version, edited by Sandy Falk. Copyright © 2024 Merck & Co., Inc., Rahway, NJ, USA and its affiliates. All rights reserved. Available at https://www.msdmanuals.com/professional.

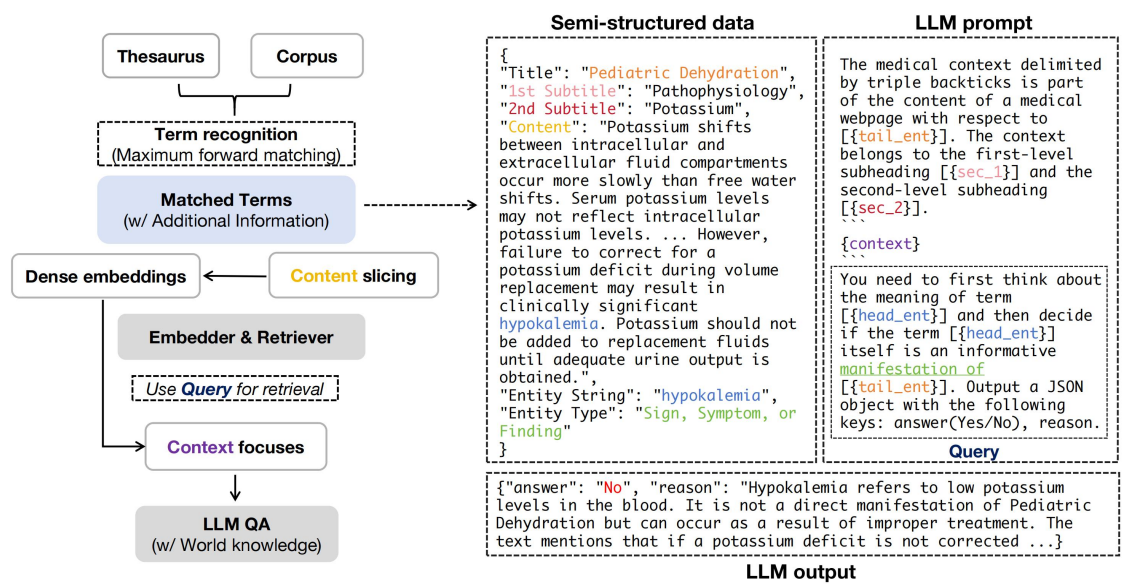

Figure 1 A high-level overview of the proposed method. Colored words denote the correspondence between the LLM prompt and the data structure. Note that the {context} part in the prompt is composed of the retrieved trunks instead of the whole context.

## General framework

The general framework is shown in **Figure 1**. Given a biomedical web article, we regard the main title as the tail entity and then treat all the biomedical terms matched on the page as head entities to check if they have a pre-defined semantic relation with the tail entity. To identify the tail entities, the preprocessed HTML data of biomedical webpages are enumeratively matched with BIOS. Due to LLMs' context length constraint, each article is sliced into text chunks, and a retrieval system is used to embed the text chunks into semantically meaningful representations and fetch the most relevant chunks to the query. Eventually, an LLM judges whether the matched head entity has a specific semantic relation (e.g., is a manifestation of) to the tail entity (i.e. the main title of the web page) and gives the reason for the decision.

## Data preprocessing

BIOS is a multilingual biomedical ontology and contains over 36 million English terms, which is ideal for extensive term-matching using dictionary-based algorithms. Based on BIOS, we extract all the biomedical terms in the web articles through maximum forward matching. Given the tremendous size of the thesaurus, the trie [28] data structure, also known as the prefix tree, is used to efficiently identify biomedical terms. Semantic types of each biomedical term from BIOS are added into the trie, assisting in selecting appropriate terms for extracting relation triplets. Specifically, the Aho-Corasick algorithm [29] is utilized to implement the maximum forward matching, in which the naive trie is equipped with additional links between internal nodes for fast transitions between matches without backtracking. After enumerating all the source texts, we obtain a series of semi-structured data with the format shown in **Figure 1**.

## Text embedding model

Oftentimes the length of the corresponding context is too long to feed in the LLM considering the constraint of predefined context length. A natural solution is to preserve a single text chunk that contains the matched head entity within a limited length. However, this may ignore the relevant context that is a bit far from the matched entity which could also benefit the relation extraction. Consequently, we chunk the whole text within one subsection with an overlap, utilize an off-the-shelf text embedding model to obtain fixed-size dense vectors for each data chunk, and retrieve the most relevant text chunks regarding the query to formulate the input context for the LLM. Given the query and each chunk, we obtain the embeddings of them, which are used to compute the cosine similarity. Then the similarity score is utilized to select the top-K similar text chunks to the query. In this work, a dense retriever *bge-large-en-v1.5* [30] is used to embed both the query and the text chunk. Specifically, *LlamaIndex* [31] is used to realize the pipeline. We first preserve the text chunk containing the matched biomedical term with a length of at least 512 words and keep a chunk size of 128 to other parts with an overlap of 32. Maximally Top-10 chunks are retrieved to form the context part in the LLM prompt.

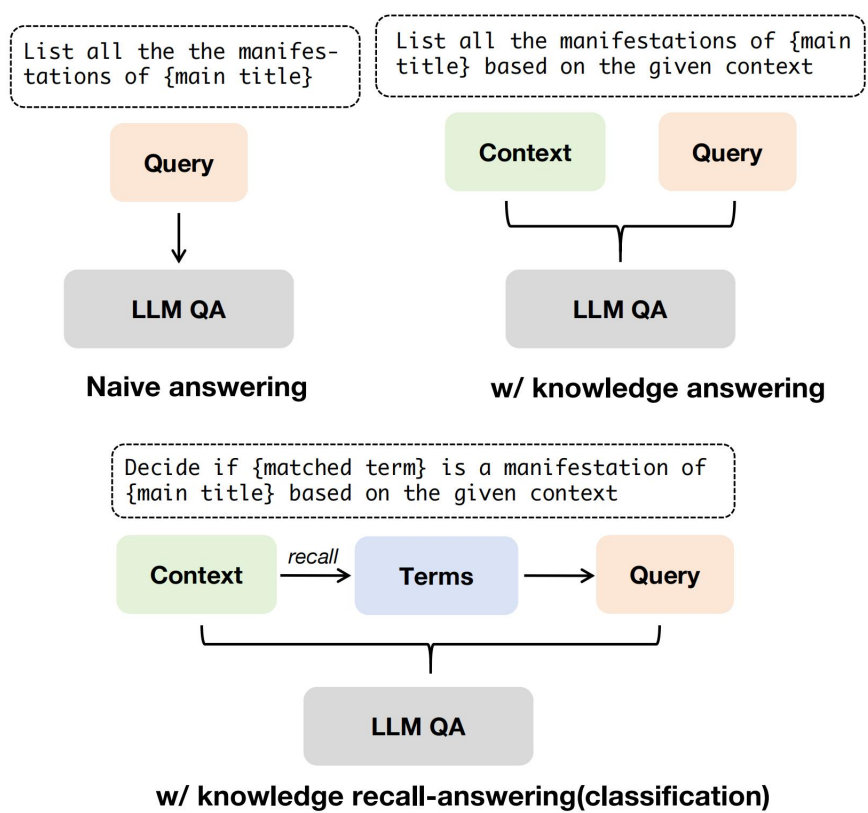

Figure 2 Different types of answering frameworks for LLMs.

**Large language models for question answering**

In this work, LLMs play a pivotal role in correctly extracting biomedical relation triplets. **Figure 2** illustrates three distinct approaches to querying a specific LLM for biomedical relation extraction. First, a naive inquiry involves directly asking the LLM about all potential entities based on a predefined semantic relation. This approach mirrors a pure open-domain question-answering problem but is prone to hallucination. Second, we insert the context into the prompt and ask the LLM to list all the relation triplets given a predefined semantic relation. While this approach is the most natural one, it may exhibit a lower recall rate, as the LLM could overlook some qualified head entities. The third way as implemented in our proposed method simplifies the relation extraction to a binary classification problem. To achieve this, we use the maximum forward matching to extract all biomedical terms in the context and use the LLM to decide whether each matched term (i.e. the head entity) presents a certain kind of biomedical relation to the main title (i.e. the tail entity).

Since the data structure is preprocessed from web HTML, it is crucial to inform the LLM explicitly that the context refers to the content corresponding to a main title on a specific web page. This specification becomes particularly important in situations where the main title (i.e. the tail entity) may not be explicitly present in the extracted context. By providing this information, the LLM is aware of the assumption that the context describes information related to the main title, especially in the context of semi-structured web articles. The LLM is required to output a JSON object with two keys: *answer* and *reason*. The *answer* key is constrained to either *Yes* or *No*, ensuring portability for large-scale extraction, while the *reason* key expounds on the decision, providing essential source attribution. Additionally, we prompt the LLM to think about the meaning of the matched entity before responding. This approach has proven to be efficient in leveraging the model's world knowledge on biomedical terms, as demonstrated in our case study.

Furthermore, we employ chain-of-thought prompting, as proposed by Wei et al. [32], to augment the Language Model Model's (LLM) answering proficiency. This approach involves presenting three exemplars that precede the specific query of interest, constituting an in-context few-shot learning strategy. These exemplars are derived from GPT-4 and meticulously verified. In practical application, the goal is to conduct high-throughput biomedical relation extraction across a vast corpus of web articles. Consequently, requiring the LLM to infer each data sample for binary question answering poses a substantial cost when utilizing API-based services such as GPT-4, despite its exceptional language generation capabilities. As a cost-effective alternative, we employ an open-source LLM named *SOLAR-0-70b-16bit*. This model is a fine-tuned version of the open-source Llama2 [33] and has demonstrated competitive performance with GPT-4 in this task.

## Benchmark construction

We evaluate four open-source LLMs: Llama2 70B, Llama2 13B, Llama2 7B, and SOLAR 70B. We compare these models with two notable commercial models, GPT-3.5 and GPT-4 with the API version of "0613". Since the allowed context length of GPT-3.5 and GPT-4 is over 4K tokens, we use the entire section as input without the slicing and retrieval steps.

We test extracting three kinds of relation types: "Is a manifestation of" for *Sign, Symptom, or Finding*, "Is a diagnostic procedure of" for *Diagnostic Procedure and Laboratory Procedure*, and "Is a treatment for" for *Therapeutic or Preventive Procedure and Chemical or Drug*. In total, 490055 potential triplets were identified from the three websites for binary classification of whether expressing the corresponding relation.

To evaluate the effectiveness of the framework and each LLM, a medical expert annotated a benchmark dataset for 155 samples randomly selected from the data. Each sample contains the classification results and the reasonings from all 6 models. To reduce annotation instability, instead of annotating independently, the expert is shown with all 6 answers from the LLMs, but the model names are masked, and the answers are randomly sorted. The expert first reviews the answers from the LLMs and then gives her classification as the gold-standard label (Yes/No, without reason for the judgement).

## Evaluation metrics

Evaluations on the expert-labeled dataset are formulated as a binary classification task. Accordingly, for each model, standard classification metrics—namely accuracy, recall, precision, and F1 score—are computed. Moreover, Cohen's kappa coefficient is computed to measure inter-LLM agreement. In instances where the LLM fails to produce the answer in a strict JSON format, the result is labeled as the opposite of the human label when evaluating classification accuracy. Similarly, in inter-LLM agreement analysis, it is labeled as the opposite of the output from GPT-4.

## Results

### Performance of LLMs on the benchmark

**Table 1** shows the classification performance of the various models according to the expert annotation. The results reveal that the open-source LLM SOLAR 70B attains the highest recall score of 0.857 and F1 score of 0.882 respectively, followed by GPT-4 with the second-highest recall score of 0.821 and F1 score of 0.881. In the meanwhile, GPT-4 achieves the highest precision of 0.950 and accuracy of 0.800, respectively. GPT-3.5 performs well in precision but relatively low in recall, and we have observed it sometimes outputs results with short reasons provided so it may be not as stable as the GPT-4 when facing knowledge-intensive tasks. The Llama2 models in general deliver performances comparable to the commercial models and the fine-tuned versions. Notably, Llama2 70B model attains the third highest F1 score of 0.827, which is competitive due to the large size. However, it's noteworthy that Llama2 13B exhibits an inferior F1 score compared to Llama2 7B due to an exceptionally low recall score. Thus the selection of the candidate model cannot be solely based on the model size when considering real practice.

Table 1 LLMs' classification performance on the expert-labeled biomedical relation extraction dataset.

|  | Accuracy | Recall | Precision | F1 |
|---|---|---|---|---|
| GPT-4 | **0.800** | 0.821 | **0.950** | 0.881 |
| GPT-3.5 | 0.658 | 0.686 | 0.914 | 0.784 |
| Llama2 70B | 0.723 | 0.736 | 0.945 | 0.827 |
| Llama2 13B | 0.503 | 0.493 | 0.920 | 0.642 |
| Llama2 7B | 0.652 | 0.693 | 0.898 | 0.782 |
| SOLAR 70B | 0.794 | **0.857** | 0.909 | **0.882** |

## Inter-LLM agreement

**Figure 2** is a heatmap representing Cohen's kappa coefficient values for predictions made by the LLMs. As is shown in the results, the cells of GPT-4, GPT-3.5, SOLAR 70B, and Llama2 70B clearly represent a high agreement group in the heatmap. SOLAR 70B and GPT-4 exhibit the highest kappa coefficient of 0.59, reflecting their close agreement in predictions. Following closely are Llama2 70B and GPT-4 (0.53), SOLAR 70B and Llama2 70B (0.51), SOLAR 70B and GPT-3.5 (0.50), and Llama2 70B and GPT-3.5 (0.49). GPT-4 and GPT-3.5 have a slightly lower agreement with the kappa coefficient of 0.45. Llama2 13B and Llama2 7B exhibit low agreement with the larger-sized models and API-based models.

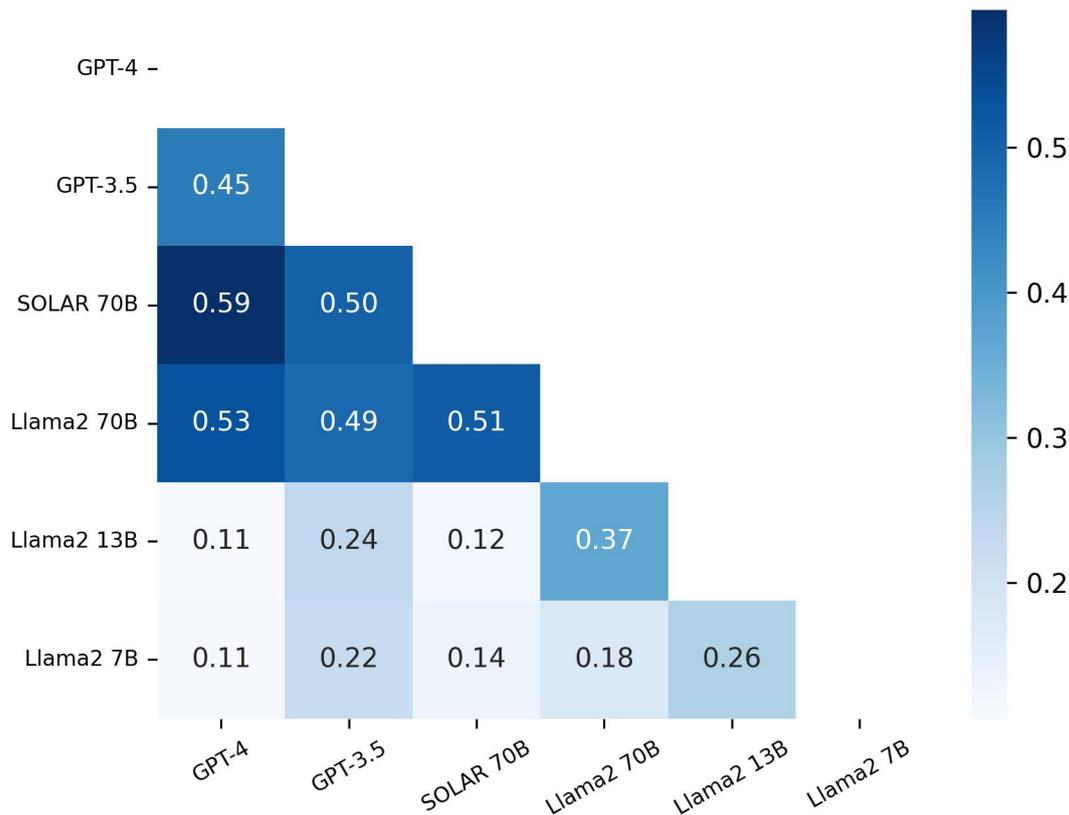

Figure 3 Heatmap illustrating inter-LLM agreement, with each matrix cell showing the correlation between predictions made by the LLMs, quantified by Cohen's kappa coefficient.

## High-throughput relation extraction in practice

Given its outstanding performance on the benchmark and strong correlation with outputs from GPT-4, SOLAR 70B is integrated into the pipeline for the biomedical relation extraction process across the three biomedical websites. Extraction statistics for the webpages are provided in **Table 2**, including information on the extracted positive semantic relation triplets for each semantic relation type and the rate of the positive classification. Cross-website triplets are deduplicated, resulting in a total number of 248659.

Table 2 Data statistics of the three selected biomedical websites.

| | Pages | Manifestation | Diagnosis | Treatment |
| --- | --- | --- | --- | --- |
| Medscape | 6216 | 80910(73.9%) | 61339(68.1%) | 118139(58.1%) |
| MSD Manual | 2288 | 9354(85.1%) | 9948(66.7%) | 19130(75.1%) |
| MedlinePlus | 1845 | 12787(84.8%) | 6021(68.6%) | 7252(61.3%) |

## Case study

Previous evaluations have primarily concentrated on binary classification outcomes, overlooking the specific reasons provided by LLMs. It is noteworthy that LLMs may occasionally yield correct binary classifications while offering flawed justifications. Oftentimes, controversial and intricate cases exist that necessitate further discussion and exploration. In **Table 3**, four illustrative cases are presented to assess the performance and limitations of LLMs. Case 1 highlights an occurrence where LLMs, including GPT-4, may occasionally over-interpret a matched term, focusing on its employment-related connotations rather than the term itself. Case 2 presents a description that nausea may appear but is rare with respect to the disease. Its classification as a manifestation is controversial and depends on the specific inclusion criteria. With the current query design, LLMs tend to lean towards denying such matched terms. Moving to Case 3, it demonstrates that LLMs such as SOLAR 70B and GPT-3.5 may misinterpret a general biomedical term, leading to incorrect decisions in question answering. However, in this case, GPT-4 does not have that issue. Finally, Case 4 says that the complete blood count (CBC) can provide help in evaluating symptoms related to the Hemolytic-uremic syndrome, which may show increased white blood cell count and decreased red blood cell count. The smaller and less advanced models struggle with correctly classifying this procedure.

Table 3 Four typical cases about performance and limitations of LLMs.

| | **Case 1** (Antisocial Personality Disorder (ASPD)) (Is a manifestation of) |
|---|---|
| **Context** | Patients with antisocial personality disorder may express their disregard for others and for the law by destroying property, harassing others, or stealing. They may deceive, exploit, con, or manipulate people to get what they want (eg, money, power, sex). ... They may change jobs with no plan for getting another. They may not seek employment when opportunities are available. They may not pay their bills, default on loans, or not pay child support. These patients are often easily provoked and physically aggressive; ... They may be charming, voluble, and verbally facile in their efforts to get what they want. |
| **Models** | **Answers** |
| GPT-4 | {"answer": "Yes", "reason": "Employment is mentioned as a factor in describing some of the behaviors of a person with Antisocial Personality Disorder. They may change jobs frequently without a plan, or may not seek employment even when opportunities are available. Hence, in the context of ASPD, employment is an informative manifestation of the disorder, indicating possible instability and irresponsibility in this area of life."} |
| SOLAR 70B | {"answer": "Yes", "reason": "Employment issues like changing jobs without having plans for another job, not seeking employment even when there are opportunities, and financial irresponsibility like not paying bills, defaulting on |

loans, or failing to pay child support can all be informative manifestations of Antisocial Personality Disorder (ASPD). People with ASPD may exhibit these behaviors because they lack concern for social norms, rules, and responsibilities towards others."}

**Case 2** (Clostridioides (formerly Clostridium) difficile–Induced Diarrhea) (Is a manifestation of)

| | |
|---|---|
| **Context** | Symptoms of C. difficile–induced diarrhea typically begin 5 to 10 days after starting antibiotics but may occur on the first day or up to 2 months later. Diarrhea may be mild and semiformed, frequent and watery, or sometimes bloody. Cramping or pain is common, but nausea and vomiting are rare. The abdomen may be slightly tender. Patients with fulminant colitis, which is characterized by severe acute inflammation of the colon and systemic toxicity, have more pain and appear very ill, with tachycardia and abdominal distention and tenderness. If colonic perforation occurs, peritoneal signs are present. |
| **Models** | **Answers** |
| GPT-4 | {"answer": "No", "reason": "Though nausea refers to the sensation of feeling sick with an inclination to vomit, in the context of C. difficile–induced diarrhea, it is mentioned that nausea and vomiting are rare symptoms."} |

**Case 3** (Plague) (Is a treatment of)

| | |
|---|---|
| **Context** | Both the prevention of symptomatic plague and treatment of plague have been recently reviewed. In adults, first line treatment is usually streptomycin or gentamicin plus doxycycline with ciprofloxacin, levofloxacin, moxifloxacin or |

chloramphenicol as secondary choices. ... Presumptive therapy (post-exposure prophylaxis) consists of a 7-day course of oral doxycycline and ciprofloxacin as first line treatment in adults. Treatment of an established case is usually for 10-14 days. In a community experiencing a pneumonic plague epidemic, individuals with a temperature of 38.5℃ or higher or newly onset cough should promptly receive parenteral antimicrobial therapy.

| Models | Answers |
|---|---|
| GPT-4 | {"answer": "No", "reason": "The term 'prophylaxis' is not a therapeutic procedure or drug. It is a medical term that refers to the prevention of disease. In the context of plague, prophylaxis may refer to various preventative measures or treatments used to prevent the disease, but it is not a specific therapeutic procedure or drug itself."} |
| SOLAR 70B | {"answer": "Yes", "reason": "Prophylaxis refers to the preventive measure taken against Plague, which includes prescribed medications like doxycycline and ciprofloxacin. These medications serve as both post-exposure prophylactic treatments and initial stages of treating established cases of Plague, making them informative therapeutic procedures/drugs for Plague."} |
| GPT-3.5 | {"answer": "Yes", "reason": "Prophylaxis is a preventive measure taken to protect against a disease. In the medical context provided, it is mentioned that consultation with an infectious diseases specialist for either prophylaxis or treatment antibiotic choices is strongly recommended for the prevention and treatment of plague. Therefore, the term prophylaxis' is an |

informative therapeutic procedure/drug for plague."}

---

**Case 4** (Hemolytic-uremic syndrome) (Is a diagnostic procedure of)

| | |
|---|---|
| **Context** | The health care provider will perform a physical exam. This may show: ||Liver or spleen swelling|| ||Nervous system changes|| Laboratory tests will show signs of hemolytic anemia and acute renal failure. Tests may include: ||Blood clotting tests (PT and PTT)|| ||Comprehensive metabolic panel may show increased levels of BUN and creatinine|| ||Complete blood count (CBC) may show increased white blood cell count and decreased red blood cell count|| ||Platelet count is usually reduced|| ||Review of a blood smear, which shows deformed and fragmented red blood cells|| ||Urinalysis may reveal blood and protein in the urine|| ||Urine protein test can show the amount of protein in the urine|| Other tests: ||Stool culture may be positive for a certain type of E coli bacteria or other bacteria|| ||Colonoscopy|| ||Kidney biopsy (in rare cases)|| |
| **Models** | **Answers** |
| GPT-4 | {"answer": "Yes", "reason": "A complete blood count (CBC) is an informative diagnostic procedure for Hemolytic-uremic syndrome. The CBC can show an increased white blood cell count and decreased red blood cell count, which are significant indicators of the disease. Additionally, it may assist in monitoring the patient's response to treatment."} |

SOLAR 70B {"answer": "Yes", "reason": "A complete blood count (CBC) is an informative diagnostic procedure for Hemolytic-uremic syndrome because it provides valuable information about the number and types of cells in the blood, including red blood cells, white blood cells, and platelets. Abnormalities in these counts can indicate the presence of the disease, specifically showing increased white blood cell count and decreased red blood cell count in the case of Hemolytic-uremic syndrome."}

Llama2 13B {"answer": "No", "reason": "The term 'complete blood count' is not a specific diagnostic procedure for Hemolytic-uremic syndrome. It is a general term referring to a laboratory test that provides information about the number and types of different cells in the blood, including red blood cells, white blood cells, and platelets. While a complete blood count may be performed as part of the diagnostic workup for Hemolytic-uremic syndrome, it alone cannot confirm the presence of the disease."}

Llama2 7B {"answer": "No", "reason": "The term 'complete blood count' is not specifically related to Hemolytic-uremic syndrome. It is a broad category of laboratory tests that measure different components of blood, including red blood cells, white blood cells, platelets, and plasma proteins. While these tests can give valuable insights into overall health status, they do not directly indicate whether someone has Hemolytic-uremic syndrome."}

## Discussion

In this study, we utilize LLMs to implement relation extraction for biomedical websites in a scalable and evidential way. We formulate the relation extraction task for the semi-structured websites as a binary classification task by recalling all the candidate terms and subsequently letting LLMs make decisions individually. Several essential ingredients contribute to the effectiveness of our proposed framework. First, we explicitly inform the LLM that the context and biomedical terms are situated on a webpage, which proved to be beneficial for the LLM to use the hierarchical sub-section information and be aware of the references in the context where the main title (i.e. the disease discussed) may not be present. Second, our framework has enhanced capability in extracting all valid relation triplets contained within the web article compared to direct querying methods. Third, given the contextual length constraint of open-source LLMs, we implement a chunk-retrieval pipeline. This approach selectively identifies the most relevant text chunks for the LLM's question-answering, optimizing the use of available information. Finally, in addition to the binary classification result, LLMs are prompted to provide the reasoning behind their decisions. This additional information is crucial for factual examination, contributing to a comprehensive understanding of the model's decision-making process.

To evaluate the performance of the proposed framework, an expert-labeled benchmark dataset is developed and LLMs with different settings are assessed with respect to classification metrics such as F1 score and agreement analyses of model predictions. Evaluation results indicate that the implemented pipeline powered by the open-source LLM shows comparable performance to the API-based services like GPT-4 and GPT-3.5. This establishes its qualification for integration into the biomedical relation extraction process. Eventually, we successfully extracted up to 248659 relation triplets across three kinds of relation types — manifestation, diagnosis, and treatment. More importantly, the proposed method exhibits scalability and can be effortlessly extended to encompass other biomedical websites and other various types of relations.

Nevertheless, it is essential to acknowledge that LLMs cannot consistently provide entirely accurate answers for every question. Case studies show that they may have over-interpreted the meaning of the matched term, resulting in the extraction of incorrect relation triplets. Models other than GPT-4 may face challenges in identifying general terms, potentially yielding uninformative relation triplets. However, this can be mitigated through post-processing by implementing keyword filtering. There exist some controversial and complex cases where the decision should be based on real needs instead of simply classifying it into Yes or No, which require more careful designs and modifications. As such, the evaluation of AI models should remain anchored in the expertise of domain-specific human experts. The relation triplets extracted by our framework serve as informative supplements to experts' handcrafted data stores, aligning with the fundamental principles of medical informatics [34].

Several directions for future improvements merit consideration. First, in this exploratory project, our focus has been exclusively on biomedical web articles where the main title pertains to the disease type. Expanding the scope to include other types, such as clinical procedures, would contribute to a more comprehensive assessment. Second, future iterations could benefit from leveraging more advanced LLMs with enhanced reading comprehension abilities, longer context lengths, and a broader factual world knowledge base. Additionally, the biomedical information retrieval model [35] is a good starting point for text chunk embedding and retrieval. Third, while we have conducted comprehensive experiments on the setting of LLM prompts, it remains uncertain whether this specific prompt format is optimal. Ongoing discussions with human experts are essential to refine and enhance the effectiveness of the LLM prompt. Last but not least, considering the LLMs' proficiency in understanding code, a potential avenue for improvement involves allowing them to directly analyze cleaned HTML text. Nevertheless, if the HTML context length exceeds the model's setting, truncations will be made and this could pose the problem of tag mismatching.

## Conclusion

High-throughput biomedical relation extraction based on semi-structured websites is a pivotal yet understudied facet of biomedical information retrieval. In this work, we propose an innovative pipeline, empowered by large language models, aimed at enabling efficient and reliable biomedical relation extraction on a large scale. The extracted relation triplets assume a central role in building diverse knowledge repositories, thereby contributing to advancements in research and applications in the fields of biomedicine and health.

## Competing interests

No competing interest is declared.

## Author contributions statement

S.Y. conceived the experiment, S.Z. conducted the experiment(s), S.Y. and S.Z. analyzed the results. S.Y. and S.Z. wrote and revised the manuscript.

## Data Availability

The input prompts and few-shot exemplars are provided in the Supplementary material.

## Acknowledgments

This work was supported by the Natural Science Foundation of China (Grant No. 12171270) and the Natural Science Foundation of Beijing Municipality (Grant No. Z190024).